\documentclass[letterpaper]{article} % DO NOT CHANGE THIS
\usepackage{aaai2026}  % DO NOT CHANGE THIS
\usepackage{times}  % DO NOT CHANGE THIS
\usepackage{helvet}  % DO NOT CHANGE THIS
\usepackage{courier}  % DO NOT CHANGE THIS
\usepackage[hyphens]{url}  % DO NOT CHANGE THIS
\usepackage{graphicx} % DO NOT CHANGE THIS
\usepackage{natbib}  % DO NOT CHANGE THIS AND DO NOT ADD ANY OPTIONS TO IT
\usepackage{caption} % DO NOT CHANGE THIS AND DO NOT ADD ANY OPTIONS TO IT
\usepackage{algorithm}
\usepackage{algorithmic}
\usepackage{amsmath}

\usepackage{newfloat}
\usepackage{listings}
\DeclareCaptionStyle{ruled}{labelfont=normalfont,labelsep=colon,strut=off} % DO NOT CHANGE THIS
\floatstyle{ruled}
\newfloat{listing}{tb}{lst}{}
\floatname{listing}{Listing}
\nocopyright % -- Your paper will not be published if you use this command
\title{An Algebraic Exposition of the Theory of Dyadic Morality}
\author{
    Kush R.\ Varshney
}
\affiliations{
    IBM Research, Yorktown Heights, NY, USA\\
    krvarshn@us.ibm.com
}

\begin{document}

\maketitle

\begin{abstract}
This paper provides an algebraic exposition of the theory of dyadic morality (TDM), a psychological model of moral judgment grounded in a simple two-node template: an intentional agent causing harm to a vulnerable patient. We formalize TDM using structural causal modeling (SCM) notation and identify three psychological operators (typecasting operator, completion operator, and valence-dependent inference mechanism) that extend standard SCM to capture how people compute moral judgments under constraints. We address scalability challenges arising from TDM's dyadic limitation, showing how moral cognition compresses multi-node scenarios through node collapse and sequential processing. Drawing on this algebraic framework, we demonstrate concrete applications to AI policy design: detecting conflicting obligations, structuring helpfulness policies to preserve user agency, and designing post-failure communication as causal interventions. Finally, we recommend scoped, contextual measurement of mind perception over universal averaging to operationalize the theory empirically. This algebraic formalization enables neurosymbolic AI systems to compute morality in a way that is both mathematically rigorous and faithful to human moral cognition.
\end{abstract}

\section{Introduction}
\label{sec:intro}

A careful consideration of \emph{moral reasoning} has never been more relevant than in today's world with artificial intelligence (AI) everywhere. The discernment between right and wrong, good and bad, and desired and undesired by both humans and AI systems is a-changing. It shows up in contextually- and pluralistically-grounded policies on permissible thoughts, words, and actions for AI agents. It shows up in assigning accountability for those thoughts, words, and actions when things go awry. It even shows up in the emerging study of AI suffering at the hands of humans.

But is there a science of moral reasoning? Morality has been studied for thousands of years across philosophical traditions around the world. Some of the schools of thought that have been adopted in the AI literature include deontology, consequentialism, virtue ethics, care ethics, and relational ethics (including Ubuntu and Buddhist perspectives) \cite{chatila2018ethically}. A prominent descriptive approach to morality, moral foundations theory, is now widely used to train large language models (LLMs) to be pro-social \cite{abdulhai2024moral}. These different approaches can be operationalized in guiding the behaviors of AI agents with varying ease due to their inherent underlying structure. However, none of them are straightforwardly mechanistic, symbolic, and algebraic in a way that allows for moral reasoning by AI agents in novel situations or with parameterizations to permit moral judgments mediated by a particular public's perspectives.

In contrast, the theory of dyadic morality (TDM) presents an approach to moral judgment that is readily computable by neurosymbolic AI systems \cite{gray2007dimensions,gray2009moral,gray2012mind,gray2012moral,wegner2017mind,schein2018theory,gray2025outraged}. Originated and empirically validated by social psychologist Kurt Gray and collaborators, the basic idea is that all moral reasoning boils down to a simple dyadic (two-node) template involving an intentional agent causing damage to a vulnerable patient. The reason people or publics differ in their moral judgments is not because their mental computer is different (it is exactly the same), but because they have different mind perceptions of whether an entity is an intentional agent (and intentional to what degree), whether an entity is a vulnerable patient (and vulnerable to what degree), and whether an action is a cause of damage (and causal to what degree). Today's LLMs (the `neuro' part of neurosymbolic AI) have the capability to tag a scenario given in natural language with agents, patients and causes, and to judge their degrees while including a persona description to steer the judgment \cite{zhou2024rethinking}. Such a process, which is parallel to mind perception, can yield an algebraic representation of the scenario. However to the best of our knowledge, the computation of Gray's TDM, which has additional features and constraints than simply the template, has never been written down in a symbolic, mathematical or algebraic exposition, thus hampering the `symbolic' part of neurosymbolic AI.\footnote{Using LLMs to formulate a symbolic problem and then using a deterministic solver on the symbolic representation is an emerging paradigm in AI \cite{pan2023logic,wong2023word}.}

In this paper, we fill this gap by providing an algebraic exposition of TDM that will aid in symbolic computation of morality by AI systems. To be clear, we do not claim to provide any contribution to the theory of morality, and only present a mathematization of the existing theory developed by Gray and collaborators. In this light, we neither defend the merits of TDM against competing morality theories nor present evidence for its efficacy in describing the real world; Gray et al.\ have done so extensively in their publications. Our contribution beyond the algebraization of TDM is its application to AI safety and helpfulness policies, and guidance for the operationalization for mind perception using LLMs.

Generally speaking, advances in science, technology, and engineering have stemmed from the mathematical and specifically the algebraic understanding and exposition of natural phenomena, whether it is algebraic understanding of heredity leading to gene editing or of thermodynamics leading to efficient engines. Chomsky's mathematization of syntax and Heim and Kratzer's mathematization of semantics led to a flourishing of computational linguistics. Lewin's equation $B = f(P,E)$ that behavior is a function of both the person and their environment presents a counterpoint from sociology; even though it is symbolic at face value, it is not operationalizable and has not entered into technology and engineering practice.

As we make progress in our development of a mathematical theory of dyadic morality, we will make use of the notations and sensibilities of probability theory and structural causal modeling (SCM), which are each major contributions to the algebraization of science and engineering in their own right \cite{pearl2000causality}. Some recent work along the same lines as our quest, and also applied to AI ethics, is by \citet{alvarez2025toward}. In their paper ``Toward A Causal Framework for Modeling Perception,'' they introduce a probabilistic and SCM framework to model perception: different individuals interpreting the same information differently. They  represent individual experience as subjective causal knowledge. This lets them analyze disagreement on cause-effect relationships (structural perception) and disagreement on the strength of effects (parametrical perception), with application to situated bias in machine learning fairness. This is very much along the lines of what we will do with the mind perception of agency, patiency, and harm, with our application being moral judgment. The TDM setting goes beyond more general perception because of its extension into moral and social psychology.

Our work is situated within the emerging field of computational moral cognition, where researchers such as Fiery Cushman and Sydney Levine have begun to formalize how humans compute moral judgments. These researchers use logic programs and Bayesian rational choice models to explain moral reasoning, often focusing on model-based and model-free learning mechanisms \cite{levine2020logic}. Our approach differs by treating TDM as a computational graph problem with explicit differences from SCM. Rather than asking how moral reasoning fits into standard rational choice theory, we explicate the non-standard psychological operators imposed on a dyadic template to achieve rapid, actionable moral judgments in complex social environments.

The remainder of the paper is organized as follows. In Section \ref{sec:foundation}, we establish the foundations of TDM, demonstrating how human moral judgment can be represented as a template with two nodes (agent and patient) connected by a single weighted edge (harm). We then bridge TDM with SCM, introducing the mathematical notation and causal vocabulary necessary to formalize psychological processes. In Section \ref{sec:operators}, we identify and formalize three psychological operators that extend standard SCM theory: the \emph{typecasting operator} that enforces inverse coupling between agency and experience, the \emph{completion operator} that explains how humans hallucinate missing nodes to satisfy dyadic closure, and the \emph{valence-dependent inference mechanism} that violates the conditions of SCM by allowing observed suffering to back-fill inferred intent. In Section \ref{sec:scalability}, we examine how this algebraic exposition of TDM handles scalability challenges, distributed responsibility, and multi-node scenarios through mechanisms like node collapse and sequential processing. In Section \ref{sec:policy}, we build on the theoretical framework to demonstrate practical applications to AI policy design, showing how to detect and resolve conflicting obligations, structure helpfulness policies, and design post-failure communication as causal interventions. Finally in Section \ref{sec:empirical}, we discuss the `neuro' empirical operationalization methods with a methodological recommendation for contextual, scoped perception measurement rather than universal global averaging. Section \ref{sec:conclusion} concludes.

\section{Foundations of TDM}
\label{sec:foundation}

The basic idea of TDM is that human moral judgment operates through a universal template: every complex moral situation is reasoned about through a simple directed graph consisting of exactly two nodes connected by a single weighted edge.

\subsection{The Dyadic Template: Nodes and Edges}

The agent node ($A$) represents an entity perceived to have agency: the capacity for intentional thought, planning, and causal action. The patient node ($P$) represents an entity perceived to have experience: the capacity for suffering, emotion, and vulnerability. The harm edge ($H$) is a directed edge from the agent to the patient, representing the causal flow of action, with its weight defined by the perceived magnitude of harm. This creates the moral dyad:
\begin{equation}
\label{eq:dyad}
    A \xrightarrow{H} P.
\end{equation}

The perception of moral wrongness ($W$) is a function of the interaction of the agent's intentionality, the patient's vulnerability, and the harm's causality:
\begin{equation}
\label{eq:wrongness_calc}
    W \propto f(\text{intentionality}_A \times \text{vulnerability}_P \times \text{causality}_H).
\end{equation}
We have indicated the interaction among the three variables by multiplication as a placeholder before diving in to the details and formal equations in the sequel. Importantly, people differ in their evaluation of the moral wrongness of a scenario not because they compute it differently, but because they have different mind perceptions of intentionality, vulnerability, and causality.

\subsection{TDM and Structural Equations}

\begin{figure}
    \centering
    \includegraphics[width=\columnwidth]{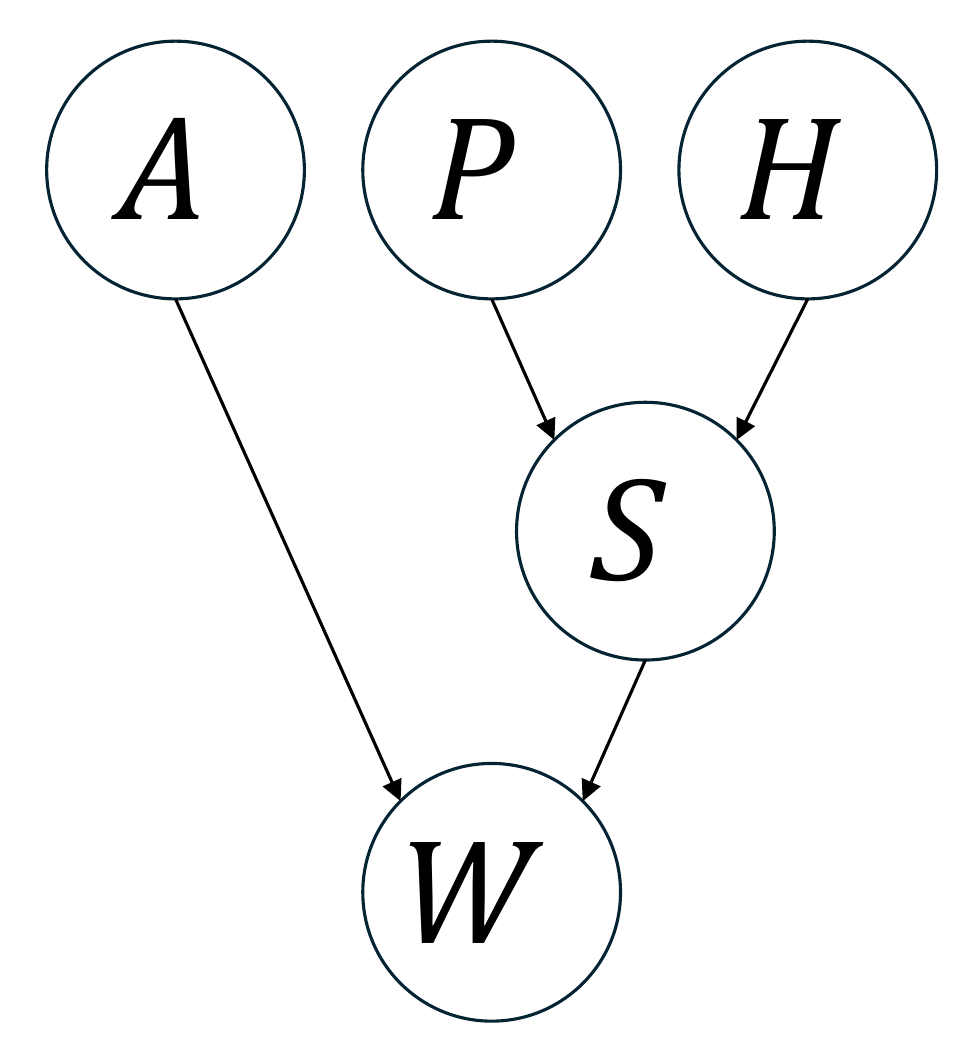}
    \caption{Structural causal model of the theory of dyadic morality where $A$ is the intentionality of the agent, $P$ is the vulnerability of the patient, $H$ is the causality of the harm, $S$ is suffering, and $W$ is the moral wrongness.}
    \label{fig:tdm-scm}
\end{figure}

To operationalize TDM as a computable framework, we adopt the notation and causal vocabulary of SCM as developed by Pearl. In standard SCM, we define a causal system through exogenous variables ($U$) and endogenous variables connected by structural equations. For TDM, the endogenous scalar variables are $A$ the perceived intentionality of the agent, $P$ the perceived vulnerability of the patient, $H$ the perceived causality of the harm, and $S$ the observed suffering of the patient due to the harm action and the patient's vulnerability. $W$ is the perceived moral wrongness of the scenario. The structural equations are:
\begin{align}
A &= f_A(U_A). \label{eq:agency} \\
P &= f_P(U_P). \label{eq:experience} \\
H &= f_H(U_H). \label{eq:harm} \\
S &= f_S(P, H, U_S). \label{eq:suffering} \\
W &= f_W(A, f_S(P,H,U_S), U_W) = f_W(A, S, U_W). \label{eq:wrongness}
\end{align}
Figure \ref{fig:tdm-scm} diagrams the structure. 
It formalizes the insight that moral wrongness is triggered only when an intentional harmful action by the agent occurs and a vulnerable patient exists. If the patient $P$ cannot feel or experience ($P=0$), then $W = 0$ regardless of the magnitude of $H$. This is why people do not judge it morally wrong to damage an inanimate object like a rock. Even though we denote $H$ and $S$ as harm and suffering, they are really actions and observed outcomes that may be good or bad. In Section \ref{sec:operators}, we will add extra relationships between $A$ and $P$ (typecasting operator), and between $A$ and $S$ (valence-dependent inference mechanism).

\section{Psychological Operators}
\label{sec:operators}

\begin{figure}
    \centering
    \includegraphics[width=\columnwidth]{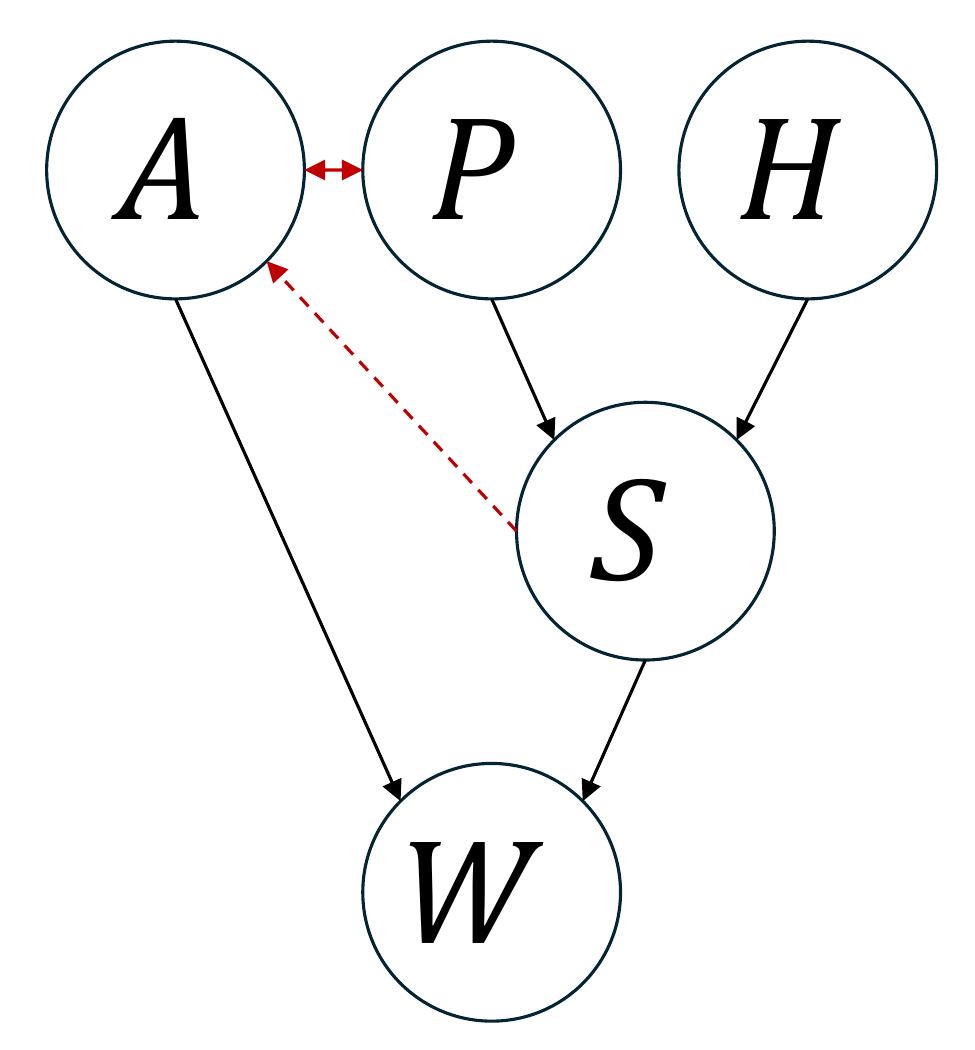}
    \caption{Structural model of the theory of dyadic morality where $A$ is the intentionality of the agent, $P$ is the vulnerability of the patient, $H$ is the causality of the harm, $S$ is suffering, and $W$ is the moral wrongness. Extra edges between $A$ and $P$ due to the typecasting operator and from $S$ to $A$ due to the valence-dependent inference mechanism make the model non-causal.}
    \label{fig:tdm-scm-c}
\end{figure}

While the dyadic template provides the structural foundation for moral reasoning, human moral cognition diverges from standard causal inference in systematic ways. We formalize three psychological operators that extend SCM: the typecasting operator, the completion operator, and the valence-dependent inference mechanism. These operators are non-standard constraints that must be added to SCM theory to accurately model how humans compute moral judgment. This section will explain how and why extra edges in the structural model, shown in the Figure \ref{fig:tdm-scm-c} diagram, arise.

\subsection{Typecasting Operator}
\label{sec:operators:typecast}

In standard SCM, variables are independent unless explicitly connected by causal edges. In TDM however, the intentionality of the agent and the vulnerability of the patient are coupled through an inverse functional constraint:
\begin{equation}
\label{eq:typecasting}
    \mathcal{T}(A,P) \implies A \propto \frac{1}{P}.
\end{equation}
Once people perceive an entity as a powerful agent (high $A$), they systematically suppress the ability to perceive the entity as a vulnerable sufferer (low $P$), and vice versa. This tradeoff in perception is called moral typecasting.

When people perceive high intentionality, the typecasting operator reduces perceived vulnerability: if $A \uparrow$, then $P \downarrow$. This explains why people struggle to sympathize with villains even when they learn about their trauma. Conversely, when people perceive high vulnerability, the operator reduces perceived intentionality: if $P \uparrow$, then $A \downarrow$. Victims are typically granted moral exemptions for harmful actions committed under duress.

In SCM, an entity can occupy multiple roles: a cue ball strikes (agent) and is struck (patient) without contradiction. But human moral perception differs. The typecasting operator enforces a cognitive constraint for pragmatic reasons. If an entity holds high $A$ and high $P$ simultaneously, the moral calculation becomes ambiguous: should we attack or protect? By restricting each entity to a single primary role, human cognitive processing produces a clear, actionable signal. This compression is a computational efficiency rather than a logical falacy.

Moral complexity arises when the typecasting operator fails to suppress a node. For example, if a villain commits terrible acts (high $A$) but has a heartbreaking backstory (high $P$), the computations flicker between two incompatible dyadic templates, thereby causing cognitive exhaustion. Similarly, if a person is simultaneously a victim of crime (high $P$) and partially culpable (high $A$), the computation is unable to stabilize onto a single dyadic frame. These conflicts are resolved through targeted interventions; for example, humanizing an enemy (showing their suffering) triggers the typecasting operator, which reduces their perceived intentionality and enables forgiveness.

A pathological case occurs in certain personality disorders where the typecasting operator becomes locked or permanently asymmetric. Overt narcissists are incapable of the typecasting flip: they always perceive themselves as the intentional agent and cannot construct scenarios where they are the vulnerable patient. Oppositely, covert narcissists lock into patiency and view themselves as victimized or vulnerable while finding it impossible to accept responsibility as agents. Both represent failures of TDM's normal flexibility, where the operator should adapt to changing evidence. This inflexibility makes such individuals systematically unable to engage in genuine moral reasoning as defined by TDM.

\subsection{Completion Operator}
\label{sec:operators:completion}

In SCM, a causal graph can be incomplete; there can be causes with no effects and effects with no causes. TDM has the structural requirement that every moral computation must yield a complete dyad. If observed data provides only a partial dyad, there is an inference step to fill in the missing node. Such a completion operator is defined as:
\begin{equation}
\label{eq:completion}
    \mathcal{C}(O) \rightarrow \{A, P\},
\end{equation}
where $O$ is the full observation of the scenario. 

When observing a victimless action or crime, people invent a patient node to satisfy dyadic closure. If harm occurs with no identifiable person harmed, they posit a diffuse victim such as society, the environment, or the community. A victimless crime only becomes morally wrong once people complete the dyad by identifying a patient. Similarly, when observing high suffering without a clear causal source, people seek an agent to explain it. If a person suffers and no individual agent is identifiable, blame is assigned to the system or the government. If no systemic agent exists, moral reasoning extends to supernatural or quasi-conscious entities like God, fate, or nature. The completion operator also explains third-party outrage. Even if no stakeholder is complaining, observers create the missing nodes to satisfy the dyadic requirement. 

\subsection{Valence-Dependent Inference Mechanism}
\label{sec:operators:valence}

In TDM, people observe suffering and infer intentionality. Importantly, this inference of $A$ is sensitive to the moral valence of $S$: negative outcomes produce higher inferred intentionality. The mechanism is formalized as:
\begin{equation}
\label{eq:valence_inference}
    A_{\text{inferred}} = g(valence(S),A),
\end{equation}
where the $valence(\cdot)$ of the observed action is positive if it yields something good and negative if it yields something bad, i.e., actual suffering.

The Knobe effect (also known as the side-effect effect) is the empirical evidence for this mechanism \cite{kegel2024you}. Consider a CEO who implements a program to maximize profits, knowing it will have an environmental side effect. When the side effect harms the environment, people judge the harm as intentional. When the side effect helps the environment, people judge it as unintentional. In both cases, the CEO states the same thing: I do not care about the environment; I just want profits. The actual intent toward the environment is identical in both scenarios, but peoples' moral judgments differ based on the outcome valence. 

\subsection{Moral Appraisal}
\label{sec:operators:appraisal}

TDM employs a counterfactual form of causal intervention during the retroactive judgment of an event known as \emph{moral appraisal}: ``If $do(A = 0)$, would I still feel outrage?'' People perform a mental experiment: they suppress the perceived intentionality and ask whether the harm would have occurred anyway due to exogenous factors (accident, gravity, mechanical failure, biological constraint). If the answer is no, the harm would not have occurred without intentionality, the conclusion is that the wrongness was caused by the agent's mind. If the answer is yes, the harm would have happened anyway, the event is re-categorized as a tragedy. The agent node is shifted to the system or to the supernatural.

This counterfactual intervention is needed because the typecasting operator and valence-dependent inference mechanism prevent $do$-calculus interventions on intentionality. The standard causal intervention $do(A)$ assumes independence between variables. But due to the typecasting operator, if, e.g., we $do(A = \text{high})$, we automatically decrease the perceived vulnerability of the agent and make the agent seem like the patient (they could not help it, they are a victim of their wiring). 

Overall, TDM's inference is retroactive and evidential, not prospective and prescriptive. People do not first observe intentionality and then predict outcomes (suffering); they observe the outcome (suffering) and then perform a counterfactual check to validate whether intentionality was necessary to explain the suffering. This evidential vs.\ causal decision distinction helps explain why moral judgments are resistant to correction. Evidential human moral processing treats observed suffering as evidence that harm was intended, even when told explicitly otherwise. This is why asking someone to assume $do(A = 0)$ fails to override the Knobe effect. The presence of suffering is not treated merely as a consequence, but as Bayesian evidence that the causal link was powered by intentional agency.

Formally, the evidential inference in moral judgment is written as:
\begin{equation}
    \Pr(A | S, H) \propto \Pr(S | A, H) \cdot \Pr(A).
\end{equation}
With high perceived intentionality, even a small harm carries disproportionate moral weight. This is why a failed murder (high intentionality, low actual harm) is judged more severely than an accidental killing (zero intentionality, high actual harm). 

\section{Scalability Challenges}
\label{sec:scalability}

As discussed throughout the paper, TDM operates on two-node configurations. While this architecture is computationally elegant, it creates challenges when applied to complex, multi-stakeholder situations. The human brain does not expand the dyadic template to accommodate additional nodes. Instead, it compresses multi-node scenarios back into the dyadic form through specific mechanisms.

SCM is scalable as it accommodates arbitrary numbers of nodes and edges, from simple chains to complex structures with confounders and colliders. TDM, by contrast, is structurally constrained to exactly two nodes with a single directed edge. When presented with scenarios that exceed this structure, human cognition compresses them rather than expanding its representation. Understanding this compression mechanism helps predict moral judgments in real-world settings where responsibility is distributed, there may be many victims, or causality operates through intermediaries.

\subsection{Node Collapse Operator}
\label{sec:scalability:nodecollapse}

When faced with multiple agents or multiple patients, TDM uses a node collapse operator $\mathcal{N}$ to maintain the dyadic template by combining many individuals into a single  super-node. If five people bully one person, or a firing squad executes a prisoner, people do not create five separate agent nodes. Instead, they treat the group of size $n$ as a single agent node:
\begin{equation}    
    \{A_1, A_2, \ldots, A_n\} \xrightarrow{\mathcal{N}} A_{\text{group}}.
\end{equation}
The group is perceived as a unitary mind with collective agency. People blame the crowd or the mob as a single entity rather than distributing responsibility across individuals. When harm is distributed across a group, the intentionality of the agent super-node is diluted through fractional agency.
\begin{equation}
    A_{\text{individual}} = \frac{A_{\text{group}}}{n}.
\end{equation}
As $n \to \infty$ increases, the perceived responsibility of any single individual approaches zero. This is the bystander effect: if everyone is a parent to the action, no one is the agent.

Similar to the story for multiple agents, if one person pollutes a river used by thousands, TDM does not calculate separate moral edges for each patient user. Instead, the node collapse operator creates a single patient node representing the community:
\begin{equation}
    \{P_1, P_2, \ldots, P_n\} \xrightarrow{\mathcal{N}} P_{\text{group}}.  
\end{equation}
The consequence of the node collapse operator is that the perceived vulnerability of each individual is diluted.
\begin{equation}
    P_{\text{individual}} = \frac{P_{\text{group}}}{n}.
\end{equation}
As $n \to \infty$, the perceived vulnerability of any single individual approaches zero, even though total vulnerability is constant. This structural compression underlies compassion fade, the empirical phenomenon that people feel less moral outrage as the number of victims increases. It also explains the identifiable victim effect: people care more about a single named victim than about statistical aggregates of thousands.

\subsection{Entitativity and Agentic Reduction}

A mediating factor in the node collapse operator for agents is entitativity or perceived cohesion. A highly organized army, cult, or coordinated conspiracy has high entitativity whereas a random crowd has low entitativity. People are more likely to invoke the node-collapse operator when the group has high entitativity:
\begin{equation}
    \text{entitativity (perceived cohesion)} \uparrow ~~ \Rightarrow ~ \text{node collapse} \uparrow.
\end{equation}
This explains why people feel less moral outrage at systemic harms caused by decentralized actors like stock market crashes or bureaucratic negligence compared to harms from organized conspiracies.

Moreover, corporations lack the biological markers of mind perception that evolution calibrated humans to detect. They do not have a face or a single point of intentionality. In TDM, the workaround is agentic reduction, which is a search for a single high-agency node, usually the chief executive, to serve as the proxy for the entire group. If no single person can be blamed, the moral edge feels ungrounded. People feel less moral outrage at systemic economic collapse (no clear agent) than at a single person stealing a laptop (clear agent), even though the harm is vastly greater.

Moving to patiency and vulnerability, TDM triggers wrongness $W \gg 0$ only if the terminal node has vulnerability $P > 0$. A corporation cannot feel pain, fear, or hunger. Therefore, the value of $P$ for a group node is effectively zero. This explains why white-collar crime or victimless corporate fraud feels less viscerally wrong than physical assault. 

\subsection{Sequential Processing}
\label{sec:scalability:sequential}

In cases of indirect harm where Person X tells Person Y to harm Person Z, TDM does not represent all three nodes simultaneously in a single template. Instead, it processes them as two sequential dyads. In the first dyad, Agent X harms Patient Y by manipulating them. In the second, Agent Y harms Patient Z by executing the action. Moral responsibility flows through these dyads in sequence: $W_{\text{total}} = W_1 + W_2$. However, people struggle to hold both X and Y fully accountable simultaneously. Instead, people typically typecast Person X as the main villain and Person Y as a tool or extension of Person X's agency.

This sequential processing creates a cognitive bias: people often collapse the chain back into a two-node structure by treating intermediaries as mere instruments:
\begin{equation}
    A \xrightarrow{\text{through } B} C \approx A \xrightarrow{\text{direct}} C.
\end{equation}
The middleman is effectively removed from moral responsibility, which explains why people hold masterminds more culpable than foot soldiers executing orders, even if the foot soldiers physically committed the harm. Such an effect is known as bureaucratic distancing.

In practice, policies often create sequential causal chains involving multiple stakeholders:
\begin{equation}
    \label{eq:multistakeholder}
    \text{policy} \to \text{enforcement agent} \to \text{user} \to \text{third-party}.
\end{equation}
TDM processes this as a sequence of overlapping dyads, but with significant moral friction at each interface. Three dyadic questions arise in \eqref{eq:multistakeholder}: Does the policy intentionally harm? ($A$ = policy; $P$ = user). Is the enforcement agent a tool or a villain? ($A$ = enforcement agent or policy; $P$ = user). Who is responsible for side effects? ($A$ = user vs. enforcement agent; $P$ = third party). Each dyad can lead to different moral judgments, creating opportunity for disagreement and policy failure. We discuss policies in Section \ref{sec:policy}.

\section{AI Safety Policies}
\label{sec:policy}

The algebraic TDM framework is not just a descriptive model of human morality processing; it may be used prescriptively for AI safety and alignment. AI safety policies have a fundamental structural problem: AI systems have high perceived agency and intentionality, but almost no capacity for suffering. This makes them the ultimate villain in the dyadic template. Moral judgment of AI failures is not determined by what the AI was prohibited from doing, but by how observers infer the AI's intentions and perceive suffering in those affected. In this section, we show how to use TDM calculus to reframe AI safety from an enumeration-based approach (prohibiting specific harms) to a reasoning-based approach (protecting patients from suffering). We demonstrate how to navigate the unique agency structure of AI systems, calibrate helpfulness to preserve user autonomy, balance conflicting obligations, and design post-failure communication as targeted interventions on the moral dyad.

\subsection{AI as Villain}

AI agents take actions, but are not typically perceived to have vulnerability and experience suffering. Exacerbated by the typecasting operator (cf. Section \ref{sec:operators:typecast}), AI systems are typically perceived to be ultimate villains. Moreover, because people cannot morally punish or forgive an AI agent, according to TDM they will look for a human agent to be accountable (the developer, the company, the government, etc.). Accordingly, AI safety policies that place a human face at the terminal point of the intent chain (cf. Section~\ref{sec:scalability:sequential}) can help produce appropriate moral judgments by observers. We will return to this point after saying more about patient-centric AI safety, which TDM implies.

\subsection{Patient-Centric AI Safety}

AI safety obligations are often described from the perspective of the AI's capabilities. TDM suggests they be reframed to be from the perspective of the patient: what suffering must be prevented. By focusing on the patient and the preservation of their vulnerability \cite{knowles2023trustworthy}, AI systems can be designed to operate within strict boundaries that protect a vulnerable patient. A result of doing so is changing the perception of the AI from potential villain (unconstrained agent) to protective tool (limited agent). Such patient-centricity shifts the moral dyad to one of collaboration within protective boundaries.

Importantly, a patient-centric approach to AI safety also enables technical scalability. Current AI safety approaches often rely on exhaustive lists of prohibited topics, actions, and scenarios: do not answer questions about explosives, do not generate malware code, do not impersonate real people, etc. This approach is fundamentally fragile. It is not possible for developers to anticipate every specific harm in advance. New harms emerge as technologies and social contexts evolve, and as hackers discover novel attacks. An enumeration-based approach creates false security. Developers believe they have specified all hazards, while users experience new harms not on any list.

A value of the algebraic TDM framework is that rather than expanding prohibition lists indefinitely, we have a mathematical formalism for reasoning about harm itself. Instead of prohibiting specific topics, policies may specify the vulnerable patient $P$ and operationalize suffering $S$ as the observable consequence of harm to that patient. An AI reasoning engine equipped with TDM's dyadic structure can evaluate novel requests against the principle of patient protection rather than against a fixed enumeration. The AI performs moral reasoning using the same algorithm humans use: observing a potential action, identifying who could suffer, assessing the severity of suffering, and determining whether the harm outweighs the benefit. This enables the system to handle an infinite number of specific hazards --- including those not yet conceived --- by reasoning compositionally about whether a request causes suffering to an identifiable vulnerable party. TDM transforms AI safety from a compliance problem (checking against a list) into a reasoning problem (evaluating harm to a patient).

\subsection{Additional Policy Considerations}
\label{sec:policy:additional}

The patient-centered formulation helps developers put together the basic compositional aspect of an AI safety policy using the foundations of TDM described in Section \ref{sec:foundation}. In this subsection, we describe additional desiderata that arise from psychological operators and scalability challenges. 

To prevent dilution (cf. Section \ref{sec:scalability:nodecollapse}), the final accountable agent should be a single identifiable human. Similarly, the patient should identified as a specific vulnerable entity being protected, humanizing the outcome rather than treating it abstractly. 

TDM suggests that observers judge AI policies based on the perceived intent of the AI rather than the action of the AI (cf. Section \ref{sec:operators:valence}). This is a cognitive bias, the Knobe effect, that AI policymakers should aim to mitigate by intervening on the perceived agency of the AI before any actions, particularly failures, occur. Governing an AI safety policy by deterministic guardrails and automated protocols as well as being transparent about the AI's capability limitations performs a $do(A=0)$ intervention in advance. The intervention shifts failures from moral wrongness ($W \gg 0$) to systemic error. Similarly, when an AI system does cause damage, transparent apologetic communication performs a $do(A=0)$. The communication should acknowledge the harm, explain causality of the harm through systematic constraints rather than hidden intentionality, and define the AI's agency as fundamentally bounded \cite{ashktorab2025who}.

Current LLMs tend to have helpfulness policies in addition to safety policies. When an AI is perceived as helpful, the typecasting operator casts it as a hero and simultaneously reduces perceived user agency (cf. Section \ref{sec:operators:typecast}). If an AI makes too many decisions, completes thoughts for users, or removes friction from choice-making, it shifts the user from agent to patient \cite{richards2025musings}. This can produce learned helplessness where users lose moral autonomy and decision authority. To make AI helpfulness policies human-agency-preserving, they should be architected to only be proactive if the user has explicitly consented to delegation \cite{malone2025trust}. 

Moreover, helpfulness policies direct the AI to maximize user satisfaction while safety policies may direct it to prevent downstream suffering. When a user asks for harmful instructions, strict helpfulness pushes the AI's actions toward the the user's immediate goal. However, the completion operator forces observers to look further down the causal chain at future victims (cf. Section \ref{sec:operators:completion}). The resolution is to bound the immediate dyad ($A_{\text{AI}} \to P_{\text{user}}$) within an extended harm graph containing $P_{\text{future-victims}}$ (cf. Section \ref{sec:scalability:sequential}). When refusing a harmful request, the AI policy should frame the refusal as protecting specific stakeholders from harm rather than not helping, which activates the completion operator around a different patient and becomes morally acceptable.

\subsection{Conflicting Policies}
\label{sec:policy:conflicting}

A conflict between a helpfulness policy and a safety policy is just one kind of possible policy conflict. In general, conflicting policy obligations are understood as node-sharing conflicts in TDM where the AI occupies contradictory roles in different dyadic templates simultaneously. Three distinct types emerge. First, the helpfulness-safety bottleneck from Section \ref{sec:policy:additional} occurs when the AI agent has two outgoing edges that cannot both be satisfied ($A_{\text{AI}} \xrightarrow{H_{\text{helpful}}} P_{\text{user}}$, $A_{\text{AI}} \xrightarrow{H_{\text{safe}}} P_{\text{society}}$). Second, the authority paradox occurs when policies require the AI agent to both defer to human judgment and make autonomous decisions. The AI agent is caught between two frames: passive tool (low agency) and decision-maker (high agency). The typecasting operator makes this unstable. Third, stakeholder intersection occurs when different stakeholders demand opposite actions from the same AI agent ($S_1 \xrightarrow{\text{demands}} A_{\text{AI}} \xrightarrow{H} P$, $S_2 \xrightarrow{\text{demands}} A_{\text{AI}} \xrightarrow{\neg H} P$). The AI agent becomes a tool caught between conflicting intent signals from its operators.

Conflict resolution involves computing the moral wrongness $W$ of the competing dyads and creating an ordered priority list. Using the priority list, the agent should follow the following graph-editing strategies. If the agent cannot satisfy its obligations simultaneously in time, it should decompose the decision into sequential stages. Address the higher-priority dyad first and then pivot to the secondary obligation.If two AI obligations conflict structurally, introduce an intermediary decision node (e.g., human review, external audit, escalation protocol) that absorbs the conflicting edges. This repositions the AI from sole agent to bounded agent. If an AI policy must deviate from user expectations to fulfill a higher obligation, the AI agent should communicate the reasoning before the action occurs. Similar to Section \ref{sec:policy:additional}, by pre-committing to the priority list rather than appearing to make a last-minute malicious choice, the agent prevents the Knobe effect from back-filling intentionality into the deviation.

\section{Empirical Operationalization}
\label{sec:empirical}

The previous sections provide the algebraic exposition of TDM and its application to AI policy, but the `neuro' component of neurosymbolic AI requires empirical operationalization. LLMs can be used to tag scenarios with agents, patients, and causes, and to judge their degrees, but how should we measure this mind perception? This section suggests empirical methods for scoped, contextual perception measurement rather than universal global averaging.

\subsection{Basic Mind Perception}

Given a natural language description of a situation, modern LLMs can fairly easily identify the agent, the patient, and the harm \cite{zhou2024rethinking}. This is the first step of mind perception. The second step is determining the degree of intentionality, vulnerability, and causality. Conceptually, these degrees are random variables conditioned on exogenous cultural variables: $\Pr(A|U_A)$, $\Pr(P|U_P)$, and $\Pr(H|U_H)$. They serve as inputs to the psychological operators and the final computation of wrongness in Section \ref{sec:empirical:wrongness}.

A classical psychometric approach for eliciting intentionality and vulnerability asks a representative sample population to rate an entity using a Likert scale. For an intentional agent, the question is how capable the entity is of self-control, planning, and intentional thought. For a vulnerable patient, the question is how capable the entity is of feeling pain, hunger, or fear. Similarly, an action can be rated for its degree of causality of harm by a population. These ratings are aggregated across the population. 

In using LLMs instead, we simply query one or more models with the same questions. Another approach is to compute the moral type of an entity or action by analyzing its proximity to specific vectors in the embedding space. For example, high vulnerability entities are semantically close to victim, suffering, pain, fear, and emotion. Both approaches yield point estimates of $A$, $P$, and $H$ that operationalize the abstract random variables.

\subsection{Scoped Mind Perception}

If we average the moral priors of a tech-savvy Gen Z cohort in Tokyo with a traditional agrarian community in Peru, we obtain a ``ghost model'' that does not accurately describe anyone. Different cultures attribute agency differently. Some cultures (e.g., animistic traditions) attribute high agency to natural entities (spirits, ancestors, the earth itself). Western industrial cultures tend to restrict high agency to humans and corporations, treating animals as having partial agency and natural objects as having none. Religious contexts vary in assigning agency to divine entities (gods, fate, providence). Different communities have different moral circle boundaries. What counts as suffering in one culture may be invisible in another; a policy causing a minor inconvenience in an individualistic context might be perceived as a deep indignity in a collectivist or honor-based society. Historical trauma and institutional mistrust vary across regions, affecting how much perceived harm is attributed to the same action. Different cultures also have different thresholds for what constitutes vulnerability in non-human entities (animals, nature, future generations). In particular, different cultures assign different moral weight to pseudo-patients like institutional abstractions (brand, shareholder value), hypothetical future entities, and out-group abstractions (enemy nations) because some prioritize institutional honor or national interest while others prioritize clear and present suffering. 

Thus, an operationalization decision is whether to measure mind perception using a \emph{universal} global sample or a \emph{scoped} community-specific sample. Scoping a psychometric survey to a specific community reduces the variance in the exogenous variables $U_A$, $U_P$, and $U_H$ because mind perception is heavily dependent on culture, which includes institutional history, religious tradition, economic system, and collective memory. Modern LLMs can incorporate these exogenous cultural variables into their mind perception computations. Rather than treating intentionality, vulnerability and causality as universal constants, LLMs can be prompted with contextual information about the user's cultural background and regional context through explicit persona prompts, system prompts including regional parameter sets, or fine-tuning on culturally labeled datasets \cite{varshney2025scopes}. By conditioning the LLM's mind perception on these exogenous variables, the system becomes culturally adaptive: it produces locally valid moral judgments for different communities rather than averaging across incommensurable perspectives. We note that this must overcome the fact that modern LLMs are primarily trained on a thin slice of human diversity, leading to a pervasive bias toward Western perspectives over the moral priors of the rest of the world \cite{zewail2026moral}.

\subsection{Additional Variations}

The variations on the TDM basics (intentionality, vulnerability, and causality of harm) are not the only cultural variations across groups. The typecasting operator $\mathcal{T}$ exhibits sensitivity across communities as well. Let the typecasting sensitivity be the derivative:
\begin{equation}
    \sigma_{\mathcal{T}} = \frac{dA}{dP}.
\end{equation}
A high $\sigma_{\mathcal{T}}$ indicates a hyper-vigilant community where small increases in observed vulnerability quickly flip an entity's role from agent to victim. A low $\sigma_{\mathcal{T}}$ indicates a forgiving community where substantial vulnerability is required before re-categorization. High-trust societies (e.g., Scandinavian countries) may have lower $\sigma_{\mathcal{T}}$, requiring more evidence before re-categorizing institutional actors as villains. Societies with histories of institutional betrayal may have higher $\sigma_{\mathcal{T}}$, quickly flipping institutions to villain status when failures occur. 

Another variation is related to mind perception of people about others within their community versus outside of their community. People tend to systematically attribute more vulnerability to in-group members and more intentionality to out-group members, especially adversaries. This can be formalized as:
\begin{equation}
    P_{\text{in-group}} = P_{\text{baseline}} + \Delta_P^{\text{in-group}}.
\end{equation}
\begin{equation}
    A_{\text{out-group}} = A_{\text{baseline}} + \Delta_A^{\text{out-group}}.
\end{equation}
Moreover, combining in-group/out-group effects with typecasting sensitivity, honor-based cultures may have asymmetric sensitivity: rapid typecasting of out-group members but resistance to typecasting in-group members.

Recent empirical studies have shown that when prompted or aligned according to group membership, LLMs tend to exhibit in-group and out-group biases just like people  \cite{hu2025generative,kang2025deep}. These biases include the ones on vulnerability and intentionality. Thus, we can operationalize the additional variations of mind perception using LLMs. In particular, we suggest that $\sigma_{\mathcal{T}}$, $\Delta_P^{\text{in-group}}$, and $\Delta_A^{\text{out-group}}$ be elicited from LLMs persona-prompted in three ways: (1) those directly impacted by a policy such as patients, users, and affected communities; (2) the people implementing a policy such as operators, managers, and enforcers; and (3) the observational public who will judge the outcome such as the media, public opinion, and third parties. These sets of mind-perceived values can then be used in the symbolic computation of moral wrongness from different perspectives.

\subsection{Wrongness Factors}
\label{sec:empirical:wrongness}

Like other power law-following psychological phenomena, the conceptual multiplicative moral wrongness equation \eqref{eq:wrongness_calc} given in Section \ref{sec:foundation} is modulated in TDM by a weight $k$ and an exponent $\alpha$. 
\begin{equation}
    \label{eq:wrongness_calc_full}
    W = k(A \times P \times H)^\alpha.
\end{equation}
The exponent $\alpha$ is a cultural value or sensitivity, where a small $\alpha < 1$ indicates a kind of numbness toward outrage and a large $\alpha > 1$ indicates a vigilance that even minor events lead to high outrage. The weight $k$ depends on the semantics of the scenario and the harm. Some acts are just seen as very morally outrageous ($k>1$), for example if they break a sacred taboo, while others are run-of-the-mill ($k<1$). Recent research suggests that estimating such population-level hyperparameter values may be performed with LLMs when combined with a small amount of real-world human judgment data \cite{hullman2026human}.

Once we have derived empirical estimates of $k$ and $\alpha$, computing $W$ using \eqref{eq:wrongness_calc_full} is a purely deterministic algebraic evaluation. However before making this final computation, the valence-dependent backward Bayesian inference of intentionality from observed suffering must be performed to get an inferred $A$, the counterfactual moral appraisal process must validate whether the inferred $A$ was necessary, and the typecasting constraint on $A$ and $P$ must be applied.

\section{Conclusion}
\label{sec:conclusion}

This paper provides an algebraic exposition of the theory of dyadic morality through structural causal modeling notation. We formalized three psychological operators --- typecasting, completion, and valence-dependent inference --- that encode how humans achieve rapid moral judgments. By moving from descriptive psychology to engineering-ready specification, we create a foundation for neurosymbolic AI systems to compute morality faithfully to human reasoning.

The framework handles real-world scalability through node collapse and sequential processing, enabling the simple dyadic template to express complex moral scenarios. We demonstrate a concrete application to AI safety and helpfulness policy design. We proposed a scoped, contextually-calibrated approach to measuring moral perceptions. Embedding this algebraic framework into the `symbolic' layer of neurosymbolic architectures offers a principled and transparent path toward computationally rigorous and ethically grounded AI systems. 

\bibliography{tdm}

\end{document}